\documentclass[preprint,authoryear]{elsarticle}
\usepackage[hidelinks]{hyperref}
\usepackage{makecell}
\usepackage{color}
\usepackage{rotating}    
\usepackage{tipa}    
\journal{Expert Systems with Applications}

\begin{document}

\begin{frontmatter}

% https://v2.overleaf.com/project/5afe9954ed7c08357b4d885d

\title{On the performance of phonetic algorithms in microtext normalization}


\author[1,2]{Yerai Doval}
%\cortext[cor]{Corresponding author: tel. +34 988 387 000, fax: +34 988 387 001}
\ead{yerai.doval@uvigo.es, yerai.doval@udc.es}
\author[1]{Manuel Vilares}
\ead{vilares@uvigo.es}
\author[2]{Jes\'us Vilares\corref{cor}}
\cortext[cor]{Corresponding author: tel. +34 981 167 000 ext. 1364, fax +34 981 167 160}
\ead{jesus.vilares@udc.es}

\address[1]{Universidade de Vigo\\Grupo COLE, Departamento de Inform\'atica, Escola Superior de Enxe\~nar\'{\i}a Inform\'atica\\Campus As Lagoas, 32004 -- Ourense (Spain)}
\address[2]{Universidade da Coru\~na\\Grupo LYS, Departamento de Computaci\'on, Facultade de Inform\'atica\\Campus de Elvi{\~n}a, 15071 -- A Coru\~na (Spain)}

\begin{abstract}

%Microtexts such as those used on microblogging platforms (e.g. Twitter), instant messaging services (e.g. WhatsApp) or traditional SMSs are usually written in a style which deviates from the standard lexical and grammatical rules of the corresponding natural language.
%This fact makes the automatic processing of these texts by traditional state--of--the--art approaches very difficult.
%Microtext normalization consists in transforming these non--standard microtexts into standard well--written texts, thus allowing us to continue to use traditional approaches.
%
%\textcolor{green}{From the perspective of intelligent and expert systems, a microtext normalization system would possess a knowledge base containing the rules to correctly handle non--standard text. 
%Given the importance of phonetic phenomena in non--standard text formation, an essential addition to this ruleset would then be the phonetic rules conforming the so--called phonetic algorithms.} 

User--generated content published on microblogging social networks constitutes a priceless source of information. However, microtexts usually deviate from the standard lexical and grammatical rules of the language, thus making its processing by traditional intelligent systems very difficult.
As an answer, microtext normalization consists in transforming those non--standard microtexts into standard well--written texts as a preprocessing step, allowing traditional approaches to continue with their usual processing.
Given the importance of phonetic phenomena in non--standard text formation, an essential element of the knowledge base of a normalizer would be the phonetic rules that encode these phenomena, which can be found in the so--called phonetic algorithms.

In this work we experiment with a wide range of phonetic algorithms for the English language. The aim of this study is to determine the best phonetic algorithms within the context of candidate generation for microtext normalization. 
In other words, we intend to find those algorithms that taking as input non--standard terms to be normalized allow us to obtain as output the smallest possible sets of normalization candidates which still contain the corresponding target standard words.
As it will be stated, the choice of the phonetic algorithm will depend heavily on the capabilities of the candidate selection mechanism which we usually find at the end of a microtext normalization pipeline.
The faster it can make the right choices among big enough sets of candidates, the more we can sacrifice on the precision of the phonetic algorithms in favour of coverage in order to increase the overall performance of the normalization system.
\end{abstract}


\end{frontmatter}

%\linenumbers

% % % % % % % % % % % % % % % % % % % % % % % % % % % % % % % % % % % %
%
% 1. INTRODUCTION
%
% % % % % % % % % % % % % % % % % % % % % % % % % % % % % % % % % % % %

\section{Introduction}
\label{sect-introduction}
With the popularization of mobile phones and Internet social networks, the use of electronic text messaging, or \emph{texting}, has reached astonishing figures such as more than 8,000 tweets produced per second.\footnote{According to \url{http://www.internetlivestats.com/one-second/} on July 2018.}
This type of communications is usually performed in real time and over platforms which impose limits on the length of the messages, as in the case of Twitter or the traditional SMS system.
Because of this, the writing style of these messages clearly differs from normal standards and phenomena such as word shortenings, contractions and abbreviations are commonly used both to gain  writing speed and circumvent length limitations.
Thus, the original well--written \emph{in--vocabulary--word} (IV) is replaced by an \emph{out--of--vocabulary--word} (OOV), as in the pairs \texttt{m8}--\texttt{mate}, \texttt{bc}--\texttt{because}, or \texttt{imo}--\texttt{in my opinion}, for example.\footnote{From now on, we will use this format OOV--IV to represent those pairs formed by an OOV and its corresponding well-written IV.}
In a similar way, phonetic-- and graphemic--based substitutions of characters are also often abused to show a more personal or customized way of writing, as in the pairs \texttt{dawg}--\texttt{dog}, \texttt{u}--\texttt{you} or \texttt{sum}--\texttt{some}, and the pairs \texttt{5o}--\texttt{so}, \texttt{0kay}--\texttt{okay} or \texttt{th3}--\texttt{the}, respectively. 
Moreover, even in the case of messaging platforms where length restrictions do not actually apply (e.g. WhatsApp), it is also common to see a writing style which tries to better reflect the feelings of the writer; for example, by using character repetitions to express emphasis, as in \texttt{noooo}--\texttt{no}, \texttt{dooooo it}--\texttt{do it} or \texttt{dammmmmn}--\texttt{damn}.
In general, these deviations from  standard writing rules are included under the general concept of \emph{texting phenomena}~\citep{thurlow2003}.

At the same time, the vast amount of data provided by social media in general and microblogging social networks in particular, constitutes an invaluable source of user--generated content of unquestionable utility for very diverse purposes: opinion mining~\citep{VilGomAlo17a}, reputation surveillance~\citep{LawGruAbr17a}, political analysis~\citep{VilTheAlo15a}, health surveillance~\citep{KarAgi18a}, crime prediction~\citep{Gerber2014a}
or disaster management~\citep{Rudraetal2016a}, to name just a few.
However, the processing and analysis of such a huge amount of data is unfeasible unless intelligent automated systems are used, which are capable of dealing with this type of content by making use of Natural Language Processing (NLP) techniques and resources~\citep{jurafsky2009}. 
Unfortunately, most of the NLP tools and resources available were originally designed to deal with standard text. Consequently, texts affected by texting phenomena cannot be reliably processed using such automated tools~\citep{Gimpel:2011:PTT:2002736.2002747,ritter2011named,foster2011news}.
As a result, two possibilities emerge in order to process this kind of texts~\citep{Eisenstein2013a}: tool adaptation, in which traditional methods which work on standard texts are reimplemented to account for the texting phenomena in non--standard texts, and text normalization.

In this work we follow the latter proposal, the so-called \emph{(micro)text normalization}, where, given a written text affected by texting phenomena, the goal is to obtain an equivalent text that better follows the writing rules of the corresponding standard language.
A well-established approach is based on performing normalization in two phases~\citep{Han:2011:LNS:2002472.2002520,saralegi2013,Schulz_etal2016a}: the first step consists of generating a set of candidate terms (i.e. possible IVs) for each input OOV; next, selection mechanisms are applied in order to rank the candidate terms and find the most likely sequence of those.
Thus, the candidate generation process is crucial in this approach; there is no need to say that if the right IV does not appear between the candidates generated, it cannot be selected in the second step.
% if we want to even consider the right IV for a given OOV.

Moreover, during the candidate generation step it is common to resort to mechanisms that exploit similarities between those words in the input and those in a dictionary of the language in order to find possible candidates.
Spell checkers such as the well--known \texttt{aspell}~\citep{atkinson2011} implement some of these mechanisms and could be used in this scenario.
Unfortunately, as can be seen in the previous texting examples, it is not uncommon for their authors to abuse the phonetic features of their language in order to shorten or customize the spelling of words.
For instance, to an English speaker, the pairs \texttt{sumone}--\texttt{someone} or \texttt{u}--\texttt{you} sound the same or quite similarly, which is also the case of the emphasis example \texttt{dooooo it} when we do not take into account the prolonged /u/.
Therefore, it would be highly  desirable to count on a mechanism which could provide us with this kind of phonetic--based matchings, and this is precisely the purpose of the so--called \emph{phonetic algorithms}~\citep{odell1918}. Given an input written word, these algorithms obtain phonetic codes that approximately indicate the way it is pronounced in a particular language. 
They can thus be used to match together words that differ in their written form but not that much in their pronunciation, as in the case of our previous examples.

However, contrary to expectations, a strict phonetic matching approach based on grapheme--to--phoneme transcription~\citep{bisani2008} using, for example, International Phonetic Alphabet (IPA) transcriptions~\cite[Ch.~7 ``Phonetics'']{jurafsky2009}, would not be useful in this context, as its output codes would be too specific for our needs. 
For instance, the processing of \texttt{beat}--\texttt{bit} or \texttt{but}--\texttt{bought} would result in slightly different codes accounting for their slightly different pronunciations 
(/bi{\textlengthmark}t/--/b{\textsci}t/ 
and 
/b{\textturnv}t/--/b{\textopeno}{\textlengthmark}t/, respectively),\footnote{This kind of word pairs are formally known as \emph{minimal pairs}.} thus preventing their matching.
So, in the context of microtext normalization, the use of a fuzzy phonetic--based matching seems a better choice.

Surprisingly, despite the importance of such phonetic processing of microtexts~\citep{Kobus_etal08b,beaufort2010,XueYinDav2011a,baldwin2013,Schulz_etal2016a} and the existence of many phonetic algorithms publicly available for the English language, we have also noticed the lack of any extensive comparison study of the performance of these algorithms for our problem domain.
In this context we have decided to conduct our own study to compare the performance of several phonetic algorithms on this task.
Our main objective is to facilitate future developers and researchers the choice of the phonetic algorithm to be used in a particular microtext normalization setup. By improving the process of normalization candidate generation, the potential effectiveness of the ulterior selection process and, consequently, that  of the whole normalization process should be notably increased. As a result, the input noise introduced into subsequent information processing systems should be greatly reduced.

Furthermore, we can model this two--step normalization procedure as an expert or intelligent system whose knowledge base specifies those transcription rules required to normalize non--standard texts.
Now, if we take into consideration the importance of phonetic--based phenomena in this domain, the need for including phonetic rules in our normalization setup should be clear.
In conclusion, a study such as the one presented here will be highly valuable when constructing a microtext normalization system, as it will guide the process of populating its knowledge base.

It should be noted that, although most of the research work on text normalization has been focusing on English~\citep{baldwin-EtAl:2015:WNUT,Han:2011:LNS:2002472.2002520,XueYinDav2011a}, there is also interest in applying it to other languages. In the European context, it is worth mentioning French~\citep{Kobus_etal08b,beaufort2010}, Dutch~\citep{Schulz_etal2016a} or Spanish, the latter being a notable case thanks to the TweetNorm Workshop~\citep{Alegria_etal2015a,Alegria_etal2013a}.
Additionally, other languages such as Chinese~\citep{WangNg2013a}, Arabic~\citep{Duwairi_etal2014a}, or even low--resource languages like Turkish~\citep{EryTor2017a} or Punjabi~\citep{Kaur_and_Singh_2015a}
% or Malay~\citep{Saloot_etal2014a} 
have been also receiving attention. 
With this in mind, the decision of using English in our experiments was mostly due to the public availability of both evaluation corpora~\citep{baldwin-EtAl:2015:WNUT} and ready--to--use implementations for a wide range of phonetic algorithms (see Section~\ref{sect-implementation-of-the-phonetic-algorithms}).

%1.3. Outline
The structure of the rest of this article is as follows. Firstly, Section~\ref{sect-phonetic-algorithms} introduces the reader to the phonetic algorithms to be analysed and how they work, while Section~\ref{sect-implementation-of-the-phonetic-algorithms} deals with their implementation. In Section~\ref{sect-evaluation}, after describing the methodology followed, the results of our experiments are presented and discussed. Next,  previously related work is introduced in Section~\ref{sect-related-work} and, finally, Section~\ref{sect-conclusions} presents our conclusions and future work.

% % % % % % % % % % % % % % % % % % % % % % % % % % % % % % % % % % % %
%
% 2. PHONETIC ALGORITHMS
%
% % % % % % % % % % % % % % % % % % % % % % % % % % % % % % % % % % % %

\section{Phonetic algorithms}
\label{sect-phonetic-algorithms}

Being able to match strings of characters which are superficially different is a key aspect of a wide range of systems such as search engines, spell checkers or, as in our case, microtext normalizers.
This usually has to do with the tendency of human actors to misspell words or, in general, deviate from  standard writing rules, both unintentionally and intentionally.
These deviations occur due to particular similarities between words, such as their spellings differing in just one character, having the same long prefix or being pronounced in almost the same way.
Thus, it would be of great interest to be able to match strings which share a particular set of features, and not (only) the totality of their constituent characters.
These types of \emph{partial} matchings are included in what is called \emph{fuzzy matching}.

Any regular spell checker, such as \texttt{aspell}~\citep{atkinson2011}, is able to perform fuzzy matching to obtain the spelling candidate corrections for an input word.
In this case, those matchings are usually determined by some distance measure being lower than a given threshold value.
The most widely used distance metric is the \emph{Levenshtein distance}, commonly known as \emph{edit distance}, which counts the number of insertions, deletions, substitutions or transpositions of characters that would need to be applied to one word to be transformed into another~\citep{levenshtein1966}.

A microtext normalization system operates in a similar way, although the candidate generation step also needs to account for texting phenomena, which cannot be handled by traditional edit--distance based approaches, as in the case of \texttt{nuff}--\texttt{enough}, \texttt{da}--\texttt{the} or \texttt{str8}--\texttt{straight}, for example.
Hence, the phonetic processing of microtexts turns out to be of key importance in order to obtain meaningful sets of normalization candidates~\citep{beaufort2010}.
This task can be accomplished through the use of so--called phonetic algorithms.

A \emph{phonetic algorithm} transforms an input written word into a phonetic code which roughly indicates the way that term is pronounced in a particular language.
It is important to highlight the \emph{approximate} nature of these codes, as its purpose is to match words with \emph{similar} pronunciations.

This work studies the most popular state--of--the--art phonetic algorithms designed for the English language in the context of the microtext normalization task.\footnote{It must be remembered that these algorithms are language--dependent.}
It is worth noting that most of them were originally designed for the task of personal--name matching, although it is fair to assume that the phonetic phenomena initially considered would also be useful in matching other types of similarly sounding words. 
Consequently, it will be interesting to analyse the performance of these algorithms in the task of microtext normalization when generating normalization candidates. 

%A brief description of each phonetic algorithm considered for the study follows, and example encodings for comparison are shown in Tables~\ref{examples12} and~\ref{examples34}.

Next, the different phonetic algorithms considered for this study are introduced. The most relevant features of each algorithm, their variations and the relations existing between them are detailed. Examples of their output encodings are also shown in Tables~\ref{examples12} and~\ref{examples34} for comparison. Notice that we have tried to be concise, keeping in mind that the objective of this work is not to study the algorithms themselves but to analyze their behaviour in the context of microtext normalization. Should the reader wish to go into detail about any particular algorithm, appropriate references have been included.

\begin{table}[tbp]
\centering
%	\begin{center}

\begin{sideways}
	\begin{footnotesize}		\begin{tabular}{lcccc}
		\hline\noalign{\smallskip}
			\textbf{algorithm} & \texttt{nuff} & \texttt{enough}  & \texttt{cntrtkxn} & \texttt{contradiction} \\ \noalign{\smallskip}\hline\noalign{\smallskip}
			\emph{Soundex} & \texttt{N100} & \texttt{E520} & \texttt{C536} & \texttt{C536} \\  
			\emph{Ref. Soundex} & \texttt{N802} & \texttt{E08040} & \texttt{C38696358} & \texttt{C308690603608} \\  
			\emph{Alpha SIS} & \texttt{02800000000000} & \texttt{12700000000000} & \small{\texttt{\makecell{07214172000000,\\06214172000000,\\07214167200000,\\06214167200000}}} & \small{\texttt{\makecell{07214171200000,\\06214171200000,\\07214161200000,\\06214161200000}}} \\  
			\emph{NYSIIS} & \texttt{NAF} & \texttt{ENAG} & \texttt{CNTRTC} & \texttt{CANTRA} \\  
			\emph{Rev. NYSIIS} & \texttt{NAF} & \texttt{ENAG} & \texttt{CNTRTCXN} & \texttt{CANTRADACTAN} \\  
			\emph{MRA} & \texttt{NF} & \texttt{ENGH} & \texttt{CNTKXN} & \texttt{CNTCTN}  \\  
			\emph{Metaphone} & \texttt{NF} & \texttt{ENKH} & \texttt{KNTRTKXN} & \texttt{KNTRTKXN} \\  
			\emph{D. Metaphone} & \texttt{NF,NF} & \texttt{ANK,ANK} & \texttt{KNTR,KNTR} & \texttt{KNTR,KNTR} \\  
			\emph{D--M Soundex} & \texttt{670000} & \texttt{065000} & \texttt{463935} & \texttt{463934} \\  
			\emph{Caverphone 1} & \texttt{NF1111} & \texttt{ANF111} & \texttt{KNTTKN} & \texttt{KNTRTK} \\  
			\emph{Caverphone 2} & \texttt{NF11111111} & \texttt{ANF1111111} & \texttt{KNTTKN1111} & \texttt{KNTRTKSN11} \\ 
			\emph{Beider--Morse} & \texttt{nuf} & \small{\texttt{\makecell{iinDg,iinog,iinug,inDg,\\inDgx,inag,inog,inogx,\\inug,inugx}}} & \small{\texttt{\makecell{kntrtgzn,kntrtkzn,\\tzntrtgzn,tzntrtkzn}}} & \tiny{\texttt{\makecell{kontradiktion,kontradiktn,\\kontraditstion,kontrodiktion,\\kontrodiktn,kontroditstion,\\kuntradiktion,kuntraditstion,\\kuntrodiktion,kuntroditstion,\\tsontradiktion,tsontraditstion,\\tsontrodiktion,tsontroditstion,\\tsuntradiktion,tsuntraditstion,\\tsuntrodiktion,tsuntroditstion}}} \\   
			\emph{F. Soundex} & \texttt{N1} & \texttt{E5} & \texttt{C536375} & \texttt{K536395} \\  
			\emph{Lein} & \texttt{N400} & \texttt{E250} & \texttt{C213} & \texttt{C213} \\  
			\emph{Onca} & \texttt{N100} & \texttt{E520} & \texttt{C536} & \texttt{C536} \\  
			\emph{Phonex} & \texttt{N1} & \texttt{A52} & \texttt{C5325} & \texttt{C5363235} \\  
			\emph{Phonix} & \texttt{N5,7} & \texttt{V5,7} & \texttt{C253632,285} & \texttt{K2536323,5} \\  
			\emph{Phonix Comm} & \texttt{N700} & \texttt{v700} & \texttt{C536} & \texttt{K536} \\   
			\emph{RogerRoot} & \texttt{02800} & \texttt{12700} & \texttt{07214} & \texttt{07214} \\  
			\emph{StatCan} & \texttt{NF} & \texttt{ENGH} & \texttt{CNTR} & \texttt{CNTR} \\  
			\emph{Eudex} & \texttt{648518346341351492} & \texttt{15564440312494426116} & \texttt{437444691622462482} & \texttt{1593606864933817373} \\ \noalign{\smallskip}\hline 
		\end{tabular}
			\end{footnotesize}
\end{sideways}
		\caption{Example encodings for each of the phonetic algorithms analysed (1): \texttt{nuff}--\texttt{enough} and \texttt{cntrtkxn}--\texttt{contradiction}}
		\label{examples12}
%	\end{center}
\end{table}

\begin{table}[tbp]
\centering
%	\begin{center}
			\begin{sideways}
	\begin{footnotesize}		\begin{tabular}{lcccc}
			\hline\noalign{\smallskip}
 		\textbf{algorithm}	& \texttt{da} & \texttt{the}  & \texttt{onez} & \texttt{ones}  \\ \noalign{\smallskip}\hline\noalign{\smallskip}
			\emph{Soundex} & \texttt{D000} & \texttt{T000} & \texttt{O520} & \texttt{O520} \\  
			\emph{Ref. Soundex}  & \texttt{D60} & \texttt{T60} & \texttt{O0805} & \texttt{O0803} \\  
			\emph{Alpha SIS} & \texttt{01000000000000} & \texttt{01000000000000} & \texttt{12000000000000} & \texttt{12000000000000} \\  
			\emph{NYSIIS} & \texttt{D} & \texttt{TH} & \texttt{ON} & \texttt{ON} \\  
			\emph{Rev. NYSIIS} & \texttt{D} & \texttt{T} & \texttt{ON} & \texttt{ON} \\  
			\emph{MRA}  & \texttt{D} & \texttt{TH} & \texttt{ONZ} & \texttt{ONS} \\  
			\emph{Metaphone} & \texttt{T} & \texttt{0} & \texttt{ONS} & \texttt{ONS} \\  
			\emph{D. Metaphone} & \texttt{T,T} & \texttt{0,T} & \texttt{ANS,ANS} & \texttt{ANS,ANS} \\  
			\emph{D--M Soundex} & \texttt{300000} & \texttt{300000} & \texttt{064000} & \texttt{064000} \\  
			\emph{Caverphone 1} & \texttt{T11111} & \texttt{T11111} & \texttt{ANS111} & \texttt{ANS111} \\  
			\emph{Caverphone 2} & \texttt{TA11111111} & \texttt{T111111111} & \texttt{ANS1111111} & \texttt{ANS1111111} \\  
			\emph{Beider--Morse} & \texttt{da,di,do} & \texttt{t,ti} & \texttt{\makecell{Ynis,Ynits,oni,oniS,\\onis,onits,unis}} & \texttt{\makecell{Ynis,oni,oniS,\\onis,unis}} \\  			
			\emph{F. Soundex} & \texttt{D} & \texttt{T} & \texttt{O59} & \texttt{O59} \\  
			\emph{Lein}  & \texttt{D000} & \texttt{T000} & \texttt{O250} & \texttt{O250} \\  
			\emph{Onca}  & \texttt{D000} & \texttt{T000} & \texttt{O500} & \texttt{O500} \\  
			\emph{Phonex} & \texttt{D} & \texttt{T} & \texttt{A52} & \texttt{A5} \\  
			\emph{Phonix} & \texttt{D,3} & \texttt{T,3} & \texttt{V5,8} & \texttt{V,58} \\  
			\emph{Phonix Comm} & \texttt{D000} & \texttt{T000} & \texttt{v800} & \texttt{v800} \\  
			\emph{RogerRoot} & \texttt{01000} & \texttt{01000} & \texttt{12000} & \texttt{12000} \\  
			\emph{StatCan} & \texttt{D} & \texttt{TH} & \texttt{ONZ} & \texttt{ONS} \\  
			\emph{Eudex} & \texttt{864691128455135232} & \texttt{1008806316530992128} & \texttt{10664523917614514324} & \texttt{10664523917614514196} \\ \noalign{\smallskip}\hline
		\end{tabular}
			\end{footnotesize}
			\end{sideways}
		\caption{Example encodings for each of the phonetic algorithms analysed (2): \texttt{da}--\texttt{the} and \texttt{onez}--\texttt{ones}.}
		\label{examples34}
%	\end{center}
\end{table}

\subsection{Soundex}
\label{algo-soundex}
Considered the first phonetic algorithm in history, the well--known and widely used \emph{Soundex} algorithm~\citep{odell1918,odell1956} mainly encodes the consonants of an input word using numerical digits, but  also encodes both consonants and vowels in the first position using that same character. 
The different digits used are related to the place of articulation of the consonant~\cite[Ch.~7: ``Phonetics'']{jurafsky2009}, so the labial consonants \texttt{b}, \texttt{f}, \texttt{p} and \texttt{v} are encoded as the number \texttt{1}, for example.
Before any other rule is applied, the algorithm checks for character sequences represented by the same number and chooses either to retain the first of those characters or the full sequence depending on other characters in the context.
Its codes have a fixed length of four characters, padded with trailing \texttt{0}'s when needed.

Soundex conforms the basis for many other modern phonetic algorithms.
These newer algorithms mostly try to address its poor precision, as in the refined version of the original algorithm (\emph{Ref. Soundex}) available in~\cite{ASF2017}, which is also tested here.
This revised version does not impose a length limit on the encodings and takes vowels more into consideration for the encoding.

\subsection{IBM Alpha Search Inquiring System}
\label{algo-alpha-sis}
Popularly known as \emph{Alpha SIS}~\citep{moore1977}, this algorithm uses two different conversion tables, one for the first characters of the input word and the other for the rest.
The encodings are conformed by a fixed number of 14 numerical digits, appending trailing \texttt{0}'s as padding for shorter words.
If two characters with the same phonetic code are adjacent, only the first one will be used, but a third character would be retained.
Alpha SIS also focuses on the encoding of consonantal sounds, although vowels are encoded if appearing at the beginning of a word. It also may return multiple alternative encodings as output.

\subsection{New York State Identification and Intelligence System}
\label{algo-nysiis}
Commonly known as \emph{NYSIIS}~\citep{taft1970}, its encoding procedure makes use of a small letter alphabet instead of numerical digits.
Its more complex ruleset with respect to the Soundex algorithm allows for the processing of notable character n--grams such as \texttt{sch} or \texttt{rd}, performing different actions depending on their context characters being vowels or not, and on characters being at the end of the word.
The first character of the word is maintained as--is while the rest of the vowels are replaced with the letter \texttt{a}.
Finally, the code is truncated to its first six characters.
			
A revised version of this algorithm (\emph{Rev. NYSIIS}), which adds new rules in order to obtain higher precision codes, is also available~\citep{kell1988}.

\subsection{Match Rating Approach}
\label{algo-mra}
Usually referred to as \emph{MRA} for short, in this case we are not only talking about a phonetic algorithm but also about a particular comparison scheme for the phonetic codes~\citep{moore1977}.
The encoding rules are quite simple: delete all vowels ---except the first one if the word begins with it---, remove character repetitions and reduce the length of the code to six by using the first and last three characters only.
However, the complexity of the system lies in the comparison rules.
For encoded strings with a length difference less than three, a matching threshold value is first calculated using a table.
Then, using a forward and backward pass over the codes of the strings to be compared, the number of distinct characters between them is obtained.
Finally, the encoded strings are considered similar depending on this number being equal or greater than the given threshold value.

\subsection{Metaphone}
\label{algo-metaphone}
Widely--used, the \emph{Metaphone} algorithm is employed in the \texttt{aspell} spell checker~\citep{philips1990}.
Its set of contextual rules maps between n--grams of characters from the source to n--grams of characters from an alphabet of sixteen consonantal symbols.
Again, the main focus of this algorithm is on encoding consonantal sounds, while vowels are included only if appearing in the first position of the word.
No limit on code length is imposed this time.

\subsection{Double Metaphone}
\label{algo-double-metaphone}
With respect to the original Metaphone, the Double Metaphone algorithm (\emph{D. Metaphone}) takes into account several spelling peculiarities from different languages, including English, and outputs two alternative encodings of the input word~\citep{philips2000}.

Although an even newer iteration in the Metaphone family exists, named Metaphone~3, unfortunately its source code is not freely available and for this reason has not been included in this study.

\subsection{Daitch-Mokotoff Soundex}
\label{algo-d-m-soundex}
The so--called Daitch--Mokotoff Soundex algorithm (\emph{D--M Soundex}) constitutes an improvement on the original Soundex seeking to improve its precision when dealing with Slavic and Yiddish surnames~\citep{mokotoff2007}.
The most notable differences with the original algorithm are the length of the codes, up to six characters long; the first character of the word being encoded as the rest, handling some specific character n--grams as a unit; and, lastly, the possibility of outputting multiple possible codes instead of a single one.

\subsection{Caverphone}
\label{algo-caverphone}
Designed by~\cite{hood2002} to match common names from New Zealand, this algorithm (\emph{Caverphone~1)} performs recursive substitutions and deletions on the original word following a set of rules mostly dealing with character n--grams.
In its last steps, it appends a padding of \texttt{1}'s to the code to finally trim its length down to six characters, which is the fixed length for all codes.

\cite{hood2004} revised his algorithm later (\emph{Caverphone 2}) by adding a few extra rules and increasing the length of the encodings to ten characters.

\subsection{Beider--Morse}
\label{algo-beider-morse}
The main goal of the \emph{Beider--Morse} algorithm~\citep{beider2008} is to reduce the large amount of false positives usually returned by Soundex.
This problem is managed by first determining the language of the input text so that a particular set of rules can be used accordingly.
If the language cannot be determined, a generic set of rules is applied instead.
A set of common rules is also applied after such language--specific or generic rules.
These common rules account for final devoicing and regressive assimilation of consonants. 
Moreover, the alphabet of this system is based on the IPA~\cite[Ch.~7 ``Phonetics'']{jurafsky2009}, which is then simplified in order to merge together symbols with very similar sounds and to make it easier to write the codes using a standard keyboard.
Finally, as with other improvements on the Soundex algorithm, it takes into account frequent character n--grams, code length is not limited, has a better support for vowels and outputs multiple possible encodings.

\subsection{Fuzzy Soundex}
\label{algo-fuzzy-soundex}
This variation of the Soundex algorithm proposed by~\cite{holmes2002} and named Fuzzy Soundex (\emph{F. Soundex}), employs two mapping tables in two subsequent stages.
In the first, a table with mappings between character n--grams is used.
Then, for the second stage, another table with mappings between individual characters and numerical digits is used to obtain the final output code, which is not restricted in length.

\subsection{Lein}
\label{algo-lein}
A simple variation of the Soundex, the \emph{Lein} algorithm differs from the original one only in its conversion table~\citep{lynch1977}.

\subsection{Onca}
\label{algo-onca}
The \emph{Onca} algorithm~\citep{gill1987,gill1993} merely consists of a two--step application of the NYSIIS and the Soundex algorithms previously described in Sections~\ref{algo-nysiis} and~\ref{algo-soundex}, respectively. According to its authors, the new algorithm overcomes the low precision of pure Soundex while retaining its 4--character format.

\subsection{Phonex}
\label{algo-phonex}
\emph{Phonex} is a Soundex--like algorithm which aims for a greater coverage of common orthographic variations~\citep{lait1996}. 
It is also influenced by the Metaphone method, hence its name.
After removing trailing \texttt{s} characters, it uses two sets of rules: one for encoding leading characters and the other for the rest of the word.
As in the original Soundex, 4--character codes are obtained as output.

\subsection{Phonix}
\label{algo-phonix}
Not to be confused with the previous Phonex algorithm, there are two variants or interpretations of the Phonix algorithm. Its original interpretation (\emph{Phonix}), described by~\cite{gadd1988,gadd1990}, specifies a set of rules to encode the input and target words as well as to obtain a matching code, called \emph{ending-sound}, for each one of them.
%It then specifies another set of matching rules regarding the word encodings and ending--sounds previously obtained, which allows for the distinction between three different categories of matches: most--likely (to be referred in this work as \emph{Phonix$_{most}$}), less--likely (\emph{Phonix$_{less}$}) and least--likely (\emph{Phonix$_{least}$}) matches.
It then specifies another set of matching rules regarding the word encodings and ending--sounds previously obtained, which allows for the distinction between three different categories of matches: most--likely, less--likely and least--likely matches (to be referred to in this work as \emph{Phonix$_{most}$}, \emph{Phonix$_{less}$} and \emph{Phonix$_{least}$}, respectively).
However, the common interpretation of the algorithm (\emph{Phonix Comm}) skips the part concerning the ending--sound and operates in a similar way to the original Soundex algorithm.

\subsection{Roger Root}
\label{algo-roger-root}
The Soundex--like \emph{Roger Root} algorithm~\citep{lynch1977} produces 5--numerical codes using two conversion tables: one for the first characters of a word and the second one for the rest.

\subsection{Census Modified Statistics Canada}
\label{algo-statcan}
Commonly known as \emph{StatCan} for short, it is a very simple phonetic algorithm which preserves the first characters, deletes the remaining vowels and \texttt{y}'s, collapses identical adjacent characters and truncates the result to four characters~\citep{lynch1977}.

\subsection{Eudex}
\label{eudex}
The \emph{Eudex} algorithm~\citep{ticki2016} encodes words in a way that exposes the differences in their pronunciations, by calculating the Hamming distances between their codes.
It returns an 8--byte array as output and makes use of four different conversion tables: two for ASCII and C1 (Latin Supplement) characters at the initial position of a word, and another two similar ones for the remaining characters.
These tables were obtained using the IPA classifications of consonants and vowels, encoding the sound articulation features of each symbol into a binary format.
This encoding has the property that similarly pronounced symbols correspond to codes with a small Hamming distance, but also highlighting the differences between them in the same way.
For this reason, its author suggests using a similarity function that matches codes with a Hamming distance below a given threshold value.

% % % % % % % % % % % % % % % % % % % % % % % % % % % % % % % % % % % %
%
% 3. IMPLEMENTATION OF THE PHONETIC ALGORITHMS
%
% % % % % % % % % % % % % % % % % % % % % % % % % % % % % % % % % % % %

\section{Implementation of the phonetic algorithms}
\label{sect-implementation-of-the-phonetic-algorithms}
In order to avoid unnecessary effort and make their future use by developers and researchers easier, instead of reimplementing from scratch the specifications for the algorithms studied, we downloaded and evaluated implementations of them that are publicly available on the Internet. 
For each downloaded implementation, its compliance with the original specification of the corresponding algorithm was tested. 
If the implementation did not comply, alternatives were sought or, when necessary, the source code was modified accordingly.\footnote{All source code produced during this work is available at \url{http://www.grupocole.org/software/VCS/phon}.} 

After having performed this selection process, these are the open source projects from which the algorithm implementations to be used were obtained:
\begin{itemize}
	\item Apache Commons Codec package~\citep{ASF2017}: in the case of the implementations of the Soundex, Refined Soundex, Daitch--Mokotoff Soundex, Beider--Morse, Caverphone 1, Caverphone 2, MRA and Double Metaphone algorithms.
	\item \texttt{talisman} package~\citep{talisman2017}: Alpha SIS, Eudex, Fuzzy Soundex, Lein, Onca, Phonex, Roger Root, StatCan, Metaphone and NYSIIS algorithms.
	\item \texttt{stringmetric} package~\citep{stringmetric2014}: Refined NYSIIS algorithm.
	\item \texttt{phonetic\_search} package~\citep{phoneticsearch2016}: Phonix algorithm.
\end{itemize}

% % % % % % % % % % % % % % % % % % % % % % % % % % % % % % % % % % % %
%
% 4. EVALUATION
%
% % % % % % % % % % % % % % % % % % % % % % % % % % % % % % % % % % % %

\section{Evaluation}
\label{sect-evaluation}
As stated before, this work intends to study the behaviour of the different phonetic algorithms available for English in the context of candidate generation for microtext normalization tasks.
The criterion assumed for this purpose states that, for each input OOV, the best algorithms should return as output the smallest possible candidate set which still contains the corresponding equivalent normalized word.

% % % % % % % % % % % % % % % % % % % % % % % % % % % % % % % % % % % %
%
% 4.1. EVALUATION CORPORA
%
% % % % % % % % % % % % % % % % % % % % % % % % % % % % % % % % % % % %

\subsection{Evaluation corpora}
Seeking reproducibility, it was decided to use the two lexical normalization dictionaries which were made publicly available in the W--NUT 2015 Shared Task~\#2~\citep{baldwin-EtAl:2015:WNUT}. These dictionaries, named \texttt{utdallas} and \texttt{unimelb}, consist of text files conformed by lists of non--standard words obtained from real--world microtexts and their corresponding standard equivalents.
It must be noted that non--standard words from these lists are affected by a wide range of texting phenomena, not limited to the phonetic phenomena in which this work is interested.
For instance, consider the graphemic substitution in \texttt{5o} (\texttt{so}) or the abbreviation \texttt{2nd} (\texttt{second}).
Despite this, the full datasets were used in the evaluation, thus resorting to error analysis to account for those instances missed by the phonetic algorithms. 

\begin{table}
	\begin{center}
		\begin{tabular}{lccc}
			\hline\noalign{\smallskip}
			\textbf{dictionary}& \textbf{\#chars/word} & \textbf{\#entries} & \textbf{\#OOV}\\ \noalign{\smallskip}\hline\noalign{\smallskip}
			\texttt{utdallas}  & 5 & 3,974   & 47\\  
			\texttt{unimelb}   & 7 & 41,181  & 96\\ \noalign{\smallskip}\hline\noalign{\smallskip} 
			\texttt{canonical} & 8 & 165,458 & -- \\ \noalign{\smallskip}\hline
			\end{tabular}
		\caption{Evaluation corpora statistics. The column \#$chars/word$ indicates for the first two rows, corresponding to the lexical normalization dictionaries, the average number of characters per non--standard word; for the last row, corresponding to the canonical dictionary, it indicates the average per standard word. Column \#$entries$ indicates the number of lines in each evaluation dataset. In the case of the first two rows, \#$OOV$ shows the number of standard words not included in the canonical lexicon used.}
		\label{stats}
	\end{center}
\end{table}

In a similar way, the canonical English dictionary made available by the W--NUT 2015 organization for the task (from now on \texttt{canonical}) was used. 
This dictionary is the list of words from which the normalization candidates are retrieved using the phonetic algorithms.
Some simple statistics of these datasets are shown in Table~\ref{stats}.

% % % % % % % % % % % % % % % % % % % % % % % % % % % % % % % % % % % %
%
% 4.2. EXPERIMENTAL METHODOLOGY
%
% % % % % % % % % % % % % % % % % % % % % % % % % % % % % % % % % % % %

\subsection{Experimental methodology}
In order to obtain the normalization candidates for each non--standard word of the evaluation corpora, the following procedure was applied:
\begin{enumerate}
	\item Before the normalization process, create the phonetic lookup dictionary of key--value pairs (phonetic code, set of words) corresponding to each phonetic algorithm.
Each pair groups all the words from the canonical dictionary with the same phonetic code for a particular algorithm.
	\item During the normalization process, and for a given phonetic algorithm, match the phonetic codes obtained from the non--standard words of the evaluation corpora with these codes (keys) of the corresponding phonetic dictionaries. 
	If a match is found, retrieve its corresponding set of words (values), which now constitute its normalization candidates.
When multiple alternative codes are available for the same non--standard input word, the final candidate set is formed by the union of those partial sets obtained with each alternative code.
\end{enumerate}

\noindent Additionally, in the case of the MRA and Phonix algorithms, their particular lookup procedures (\emph{MRA}$_{custom}$, \emph{Phonix}$_{most}$, \emph{Phonix}$_{less}$ and \emph{Phonix}$_{least}$, respectively), previously explained in Section~\ref{sect-phonetic-algorithms}, were also considered.
Similarly, multiple results were obtained for the Eudex algorithm by varying the Hamming distance threshold value: \emph{Eudex}, \emph{Eudex$_{5}$}, \emph{Eudex}$_{10}$ and \emph{Eudex}$_{15}$ for threshold values $0$ (i.e. perfect matching), $5$, $10$ and $15$, respectively.
In these cases, both the results corresponding to the general and specialized lookup procedures were obtained.\footnote{For the MRA and Eudex algorithms, it is worth mentioning the relatively high computational cost of their particular lookup procedures. Under their current implementations, they may be suitable for quick checks between pairs of words but not for the extensive dictionary--wide checks required in the current context.}

Regarding the original Phonix algorithm in particular, three different sets of results were distinguished based on the likelihood of the candidates indicated by the lookup procedure: $(1)$ one set containing the most likely candidates (\emph{Phonix}$_{most}$); $(2)$ one including the most and less likely candidates (\emph{Phonix}$_{less}$) and $(3)$ one which also adds the least likely candidates (\emph{Phonix}$_{least}$).
Moreover, results obtained using the alternative interpretation of this algorithm are also included (\emph{Phonix}$_{Comm}$).

Precision, recall and F1 metrics, over a list of words from a particular evaluation dataset, are used to measure the performance of the phonetic algorithms. 
In the present application context, \emph{precision} (P) is defined as the mean of the ratio of correct candidates\footnote{In this case, the number of correct candidates will be either 1 or 0.} over the total number of candidates retrieved for each word, all of this over the total number of words:

\begin{equation}
	P = \frac{\sum_{word}\frac{|hits_{word}|}{|candidates_{word}|}}{|words|} 
\end{equation}

\noindent In the case of \emph{recall} (R), it is calculated as the number of times when the correct candidate was among the set of normalization candidates retrieved for each word over the total number of words:

\begin{equation}
	R = \frac{|hits|}{|words|}
\end{equation}

\noindent Finally, \emph{F1} score is defined in the usual way by aggregating precision and recall:

\begin{equation}
	F1 = 2 \cdot \frac{P \cdot R}{P + R}
\end{equation}

% % % % % % % % % % % % % % % % % % % % % % % % % % % % % % % % % % % %
%
% 4.3. RESULTS AND DISCUSSION
%
% % % % % % % % % % % % % % % % % % % % % % % % % % % % % % % % % % % %

\subsection{Results and discussion}

%Tables~\ref{results-utdallas-F1} and~\ref{results-unimelb-F1} show the results obtained for each algorithm, sorting them by their corresponding F1 scores, while Tables~\ref{results-utdallas-recall} and~\ref{results-unimelb-recall} show them sorted by their recall values instead.

\begin{table}[tb]
\centering
%	\begin{center}
\begin{footnotesize}
\begin{tabular}{lcrccc} 
	\hline\noalign{\smallskip}
	\textbf{algorithm} & \textbf{\#hits} & \textbf{avg \#cands} & \textbf{P} & \textbf{R} & \textbf{F1} \\
	\noalign{\smallskip}\hline\noalign{\smallskip}
	\emph{Eudex} &     1,376     &              5.641              &   0.117    &   0.346    &    0.175    \\
	\emph{MRA}                                                   &     1,498     &              9.165              &   0.113    &   0.376    &    0.174    \\
	\emph{Metaphone}                                             &     2,071     &             26.232              &   0.078    &   0.521    &    0.137    \\
	\emph{Beider--Morse}                                         &     1,636     &             25.514              &   0.080    &   0.411    &    0.134    \\
	\emph{StatCan}                                               &     2,225     &             21.415              &   0.070    &   0.559    &    0.124    \\
	\emph{Ref. Soundex}                                          &     1,569     &             15.378              &   0.073    &   0.394    &    0.123    \\
	\emph{Eudex}$_{5}$                                           &     1,800     &             94.125              &   0.047    &   0.452    &    0.085    \\
	\emph{Rev. NYSIIS}                                           &     1,416     &             27.404              &   0.047    &   0.356    &    0.083    \\
	\emph{Phonix}$_{most}$                                       &     1,681     &             36.503              &   0.044    &   0.422    &    0.080    \\
	\emph{F. Soundex}                                            &     2,160      &             41.765              &   0.043    &   0.543    &    0.080    \\
	\emph{NYSIIS}                                                &     1,410      &             17.565              &   0.043    &   0.354    &    0.078    \\
	\emph{Caverphone 2}                                          &     2,039      &             69.022              &   0.039    &   0.513    &    0.073    \\
	\emph{Caverphone 1}                                          &     2,246      &             98.240              &   0.031    &   0.565    &    0.060    \\
	\emph{Eudex}$_{10}$                                          &     2,076      &             310.535             &   0.025    &   0.522    &    0.048    \\
	\emph{D--M Soundex}                                          &     2,154      &             96.137              &   0.024    &   0.542    &    0.046    \\
	\emph{Alpha SIS}                                             &     2,359      &             241.792             &   0.022    &   0.593    &    0.042    \\
	\emph{Phonix}$_{less}$                                       &     2,191      &             141.575             &   0.020    &   0.551    &    0.040    \\
	\emph{Eudex}$_{15}$                                          &     2,199      &             551.689             &   0.016    &   0.553    &    0.032    \\
	\emph{Roger Root}                                            &     2,516      &             133.588             &   0.016    &   0.633    &    0.032    \\
	\emph{Soundex}                                               &     2,469      &             80.570              &   0.015    &   0.621    &    0.030    \\
	\emph{D. Metaphone}                                          &     2,395      &             84.389              &   0.014    &   0.602    &    0.028    \\
	\emph{Lein}                                                  &     2,471      &             101.642             &   0.013    &   0.621    &    0.026    \\
	\emph{Phonix Comm}                                           &     2,453      &             194.972             &   0.011    &   0.617    &    0.023    \\
	\emph{Onca}                                                  &     2,396      &             109.237             &   0.010    &   0.602    &    0.020    \\
	\emph{Phonex}                                                &     2,656      &             266.348             &   0.009    &   0.668    &    0.019    \\
	\emph{Phonix}$_{least}$                                      &     2,332      &             611.050             &   0.006    &   0.586    &    0.013    \\
	\emph{MRA}$_{custom}$                                        &     3,695      &            15,270.327            &   0.000    &   0.929    &    0.000    \\
	\noalign{\smallskip}\hline                                   
\end{tabular}
\end{footnotesize}
\caption{Results for the \texttt{utdallas} dictionary, ranked by F1. For each phonetic algorithm, the second column (\emph{\#hits}) indicates the number of instances from the corpora for which the algorithm could provide the correct answer. The average number of normalization candidates returned for each OOV using that algorithm is shown in the third column (\emph{avg \#cands}). The rest of the columns contain the values obtained for the evaluation metrics used; from left to right: precision (\emph{P}), recall (\emph{R}) and F1 score (\emph{F1}).}
\label{results-utdallas-F1}
\end{table}

\begin{table}[tb]
\centering
%	\begin{center}
		\begin{footnotesize}
\begin{tabular}{lcrccc} 
			\hline
			\noalign{\smallskip}
	\textbf{algorithm} & \textbf{\#hits} & \textbf{avg \#cands} & \textbf{P} & \textbf{R} & \textbf{F1} \\
			\noalign{\smallskip}\hline\noalign{\smallskip}
			\emph{MRA} &     17,485     &              5.963              &   0.171    &   0.424    &    0.244    \\
			\emph{Eudex}                                                 &     16,150     &              3.682              &   0.174    &   0.392    &    0.241    \\
			\emph{Metaphone}                                             &     22,448     &             16.640              &   0.136    &   0.545    &    0.218    \\
			\emph{Ref. Soundex}                                          &     18,693     &             10.464              &   0.126    &   0.453    &    0.197    \\
			\emph{Beider--Morse}                                         &     18,307     &             13.792              &   0.123    &   0.444    &    0.193    \\
			\emph{Rev. NYSIIS}                                           &     17,844     &             16.149              &   0.091    &   0.433    &    0.150    \\
			\emph{F. Soundex}                                            &     23,694     &             28.393              &   0.083    &   0.575    &    0.145    \\
			\emph{Eudex}$_{5}$                                           &     18,498     &             42.203              &   0.085    &   0.449    &    0.143    \\
			\emph{StatCan}                                               &     28,632     &             29.902              &   0.076    &   0.695    &    0.138    \\
			\emph{Phonix}$_{most}$                                       &     19,039     &             23.382              &   0.081    &   0.462    &    0.137    \\
			\emph{Caverphone 2}                                          &     21,580     &             43.374              &   0.075    &   0.524    &    0.132    \\
			\emph{NYSIIS}                                                &     21,094     &             17.039              &   0.066    &   0.512    &    0.118    \\
			\emph{Caverphone 1}                                          &     24,227     &             61.604              &   0.062    &   0.588    &    0.112    \\
			\emph{Eudex}$_{10}$                                          &     19,988     &             150.981             &   0.054    &   0.485    &    0.097    \\
			\emph{Alpha SIS}                                             &     26,064     &             145.874             &   0.048    &   0.632    &    0.089    \\
			\emph{D--M Soundex}                                          &     24,470     &             64.662              &   0.045    &   0.594    &    0.084    \\
			\emph{Phonix}$_{less}$                                       &     23,093     &             83.416              &   0.041    &   0.560    &    0.077    \\
			\emph{Eudex}$_{15}$                                          &     20,892     &             290.397             &   0.038    &   0.507    &    0.071    \\
			\emph{Roger Root}                                            &     28,928     &             98.075              &   0.028    &   0.702    &    0.054    \\
			\emph{Phonex}                                                &     28,048     &             200.475             &   0.020    &   0.681    &    0.040    \\
			\emph{D. Metaphone}                                          &     26,253     &             89.951              &   0.020    &   0.637    &    0.039    \\
			\emph{Soundex}                                               &     29,557     &             94.946              &   0.018    &   0.717    &    0.035    \\
			\emph{Phonix Comm}                                           &     26,937     &             134.754             &   0.018    &   0.654    &    0.035    \\
			\emph{Phonix}$_{least}$                                      &     24,434     &             387.919             &   0.017    &   0.593    &    0.034    \\
			\emph{Onca}                                                  &     28,601     &             109.179             &   0.014    &   0.694    &    0.028    \\
			\emph{Lein}                                                  &     29,633     &             134.894             &   0.012    &   0.719    &    0.025    \\
			\emph{MRA}$_{custom}$                                        &     38,768     &            20,767.367            & 6.873e-05  &   0.941    &    0.000    \\
				\noalign{\smallskip}\hline
		\end{tabular}
		\end{footnotesize}
		\caption{Results for the \texttt{unimelb} dictionary, ranked by F1.}
		\label{results-unimelb-F1}
%	\end{center}
\end{table}

%As can be seen in Tables~\ref{results-utdallas-F1} and~\ref{results-unimelb-F1}, the F1 scores obtained are quite low, mainly due to the low precision figures obtained in most of the cases. Accordingly, those algorithms with the highest precision scores end up at the top of this ranking.

Tables~\ref{results-utdallas-F1} and~\ref{results-unimelb-F1} show the results obtained for each algorithm, sorting them by their corresponding F1 scores. As it can be seen, the scores obtained are quite low, mainly due to the low precision figures obtained in most of the cases. Accordingly, those algorithms with the highest precision scores end up at the top of this ranking, with \emph{Eudex}, \emph{MRA} and \emph{Metaphone} outperforming the rest of them.

It is interesting to note that the performance of the MRA algorithm decreases ostensibly when used with its particular lookup procedure (\emph{MRA}$_{custom}$).
This procedure tries to enlarge the matching window of its otherwise high--precision codes.
Thus, it is reasonable to assume that using a large canonical dictionary, as in this case, renders this procedure of little use, as it was designed to work with much smaller lists.
Moreover, \emph{Phonix}$_{least}$, that is, Phonix using least--likely matches, suffers from the same problem.

\begin{table}[tb]
\centering
%	\begin{center}
	\begin{footnotesize}
\begin{tabular}{lcrccc} 
			\hline
			\noalign{\smallskip}
	\textbf{algorithm} & \textbf{\#hits} & \textbf{avg \#cands} & \textbf{P} & \textbf{R} & \textbf{F1} \\
			\noalign{\smallskip}\hline\noalign{\smallskip}
			\emph{MRA}$_{custom}$ &     3,695      &            15,270.327            &   0.000    &   0.929    &    0.000    \\
			\emph{Phonex}                                                           &     2,656      &             266.348             &   0.009    &   0.668    &    0.019    \\
			\emph{Roger Root}                                                       &     2,516      &             133.588             &   0.016    &   0.633    &    0.032    \\
			\emph{Lein}                                                             &     2,471      &             101.642             &   0.013    &   0.621    &    0.026    \\
			\emph{Soundex}                                                          &     2,469      &             80.570              &   0.015    &   0.621    &    0.030    \\
			\emph{Phonix Comm}                                                      &     2,453      &             194.972             &   0.011    &   0.617    &    0.023    \\
			\emph{Onca}                                                             &     2,396      &             109.237             &   0.010    &   0.602    &    0.020    \\
			\emph{D. Metaphone}                                                     &     2,395      &             84.389              &   0.014    &   0.602    &    0.028    \\
			\emph{Alpha SIS}                                                        &     2,359      &             241.792             &   0.022    &   0.593    &    0.042    \\
			\emph{Phonix}$_{least}$                                                 &     2,332      &             611.050             &   0.006    &   0.586    &    0.013    \\
			\emph{Caverphone 1}                                                     &     2,246      &             98.240              &   0.031    &   0.565    &    0.060    \\
			\emph{StatCan}                                                          &     2,225      &             21.415              &   0.070    &   0.559    &    0.124    \\
			\emph{Eudex}$_{15}$                                                     &     2,199      &             551.689             &   0.016    &   0.553    &    0.032    \\
			\emph{Phonix}$_{less}$                                                  &     2,191      &             141.575             &   0.020    &   0.551    &    0.040    \\
			\emph{F. Soundex}                                                       &     2,160      &             41.765              &   0.043    &   0.543    &    0.080    \\
			\emph{D--M Soundex}                                                     &     2,154      &             96.137              &   0.024    &   0.542    &    0.046    \\
			\emph{Eudex}$_{10}$                                                     &     2,076      &             310.535             &   0.025    &   0.522    &    0.048    \\
			\emph{Metaphone}                                                        &     2,071      &             26.232              &   0.078    &   0.521    &    0.137    \\
			\emph{Caverphone 2}                                                     &     2,039      &             69.022              &   0.039    &   0.513    &    0.073    \\
			\emph{Eudex}$_{5}$                                                      &     1,800      &             94.125              &   0.047    &   0.452    &    0.085    \\
			\emph{Phonix}$_{most}$                                                  &     1,681      &             36.503              &   0.044    &   0.422    &    0.080    \\
			\emph{Beider--Morse}                                                    &     1,636      &             25.514              &   0.080    &   0.411    &    0.134    \\
			\emph{Ref. Soundex}                                                     &     1,569      &             15.378              &   0.073    &   0.394    &    0.123    \\
			\emph{MRA}                                                              &     1,498      &              9.165              &   0.113    &   0.376    &    0.174    \\
			\emph{Rev. NYSIIS}                                                      &     1,416      &             27.404              &   0.047    &   0.356    &    0.083    \\
			\emph{NYSIIS}                                                           &     1,410      &             17.565              &   0.043    &   0.354    &    0.078    \\
			\emph{Eudex}                                                            &     1,376      &              5.641              &   0.117    &   0.346    &    0.175    \\
				\noalign{\smallskip}\hline             
		\end{tabular}
		\end{footnotesize}
				\caption{Results for the \texttt{utdallas} dictionary, this time ranked by recall.}
		\label{results-utdallas-recall}
%	\end{center}
\end{table}

\begin{table}[tb]
\centering
%	\begin{center}
	\begin{footnotesize}
\begin{tabular}{lcrccc} 
			\hline
			\noalign{\smallskip}
	\textbf{algorithm} & \textbf{\#hits} & \textbf{avg \#cands} & \textbf{P} & \textbf{R} & \textbf{F1} \\
			\noalign{\smallskip}\hline\noalign{\smallskip}
			\emph{MRA}$_{custom}$ &     38,768     &            20,767.367            & 6.873e-05  &   0.941    &    0.000    \\
			\emph{Lein}                                                             &     29,633     &             134.894             &   0.012    &   0.719    &    0.025    \\
			\emph{Soundex}                                                          &     29,557     &             94.946              &   0.018    &   0.717    &    0.035    \\
			\emph{Roger Root}                                                       &     28,928     &             98.075              &   0.028    &   0.702    &    0.054    \\
			\emph{StatCan}                                                          &     28,632     &             29.902              &   0.076    &   0.695    &    0.138    \\
			\emph{Onca}                                                             &     28,601     &             109.179             &   0.014    &   0.694    &    0.028    \\
			\emph{Phonex}                                                           &     28,048     &             200.475             &   0.020    &   0.681    &    0.040    \\
			\emph{Phonix Comm}                                                      &     26,937     &             134.754             &   0.018    &   0.654    &    0.035    \\
			\emph{D. Metaphone}                                                     &     26,253     &             89.951              &   0.020    &   0.637    &    0.039    \\
			\emph{Alpha SIS}                                                        &     26,064     &             145.874             &   0.048    &   0.632    &    0.089    \\
			\emph{D--M Soundex}                                                     &     24,470     &             64.662              &   0.045    &   0.594    &    0.084    \\
			\emph{Phonix}$_{least}$                                                 &     24,434     &             387.919             &   0.017    &   0.593    &    0.034    \\
			\emph{Caverphone 1}                                                     &     24,227     &             61.604              &   0.062    &   0.588    &    0.112    \\
			\emph{F. Soundex}                                                       &     23,694     &             28.393              &   0.083    &   0.575    &    0.145    \\
			\emph{Phonix}$_{less}$                                                  &     23,093     &             83.416              &   0.041    &   0.560    &    0.077    \\
			\emph{Metaphone}                                                        &     22,448     &             16.640              &   0.136    &   0.545    &    0.218    \\
			\emph{Caverphone 2}                                                     &     21,580     &             43.374              &   0.075    &   0.524    &    0.132    \\
			\emph{NYSIIS}                                                           &     21,094     &             17.039              &   0.066    &   0.512    &    0.118    \\
			\emph{Eudex}$_{15}$                                                     &     20,892     &             290.397             &   0.038    &   0.507    &    0.071    \\
			\emph{Eudex}$_{10}$                                                     &     19,988     &             150.981             &   0.054    &   0.485    &    0.097    \\
			\emph{Phonix}$_{most}$                                                  &     19,039     &             23.382              &   0.081    &   0.462    &    0.137    \\
			\emph{Ref. Soundex}                                                     &     18,693     &             10.464              &   0.126    &   0.453    &    0.197    \\
			\emph{Eudex}$_{5}$                                                      &     18,498     &             42.203              &   0.085    &   0.449    &    0.143    \\
			\emph{Beider--Morse}                                                    &     18,307     &             13.792              &   0.123    &   0.444    &    0.193    \\
			\emph{Rev. NYSIIS}                                                      &     17,844     &             16.149              &   0.091    &   0.433    &    0.150    \\
			\emph{MRA}                                                              &     17,485     &              5.963              &   0.171    &   0.424    &    0.244    \\
			\emph{Eudex}                                                            &     16,150     &              3.682              &   0.174    &   0.392    &    0.241    \\
				\noalign{\smallskip}\hline
		\end{tabular}
		\end{footnotesize}
				\caption{Results for the \texttt{unimelb} dictionary, this time ranked by recall.}
		\label{results-unimelb-recall}
%	\end{center}
\end{table}

At this point, it should be noted that, when developing a microtext normalization system, it may be interesting to gain some recall while sacrificing some precision in exchange. This is due to the fact that a low recall tends to impose an upper limit on the overall performance of the system: there is no way of selecting the right IV unless it appears among the generated candidates.
% precision in exchange, since a low recall tends to impose an upper limit on the overall performance of the system.
Moreover, it is also reasonable to assume that the candidate generation step will be connected to a capable candidate selection step afterwards.
Taking this into account, Tables~\ref{results-utdallas-recall} and~\ref{results-unimelb-recall} show again the results obtained in our experiments, but this time ranked by their recall figures. The resulting rankings are now very different, providing us with new insights about the performance of the analyzed phonetic algorithms.
%scores. As can be seen, the resulting rank is now very different.

Overall, these results indicate that the \emph{StatCan}, \emph{Metaphone}, %\emph{F.Soundex}, 
\emph{Soundex} or \emph{Roger Root} algorithms would be good choices for candidate generation in a microtext normalization system, depending on the level of compromise sought between precision and recall.
This way, the \emph{Metaphone} algorithm shines when precision is needed, as it is among the top--three algorithms in the F1 classification (see Tables~\ref{results-utdallas-F1} and~\ref{results-unimelb-F1}) while it does not fall to the bottom of the recall rankings (Tables~\ref{results-utdallas-recall} and~\ref{results-unimelb-recall}) as in the case of \emph{MRA} or \emph{Eudex}, the other two algorithms with the highest F1. On the other hand, \emph{Soundex} and \emph{Roger Root} stand as good choices if we want to maximize recall while not hurting precision in excess, as in the case of the \emph{Lein} algorithm, and after having dismissed $MRA_{custom}$ for the reasons given above. Finally, \emph{Statcan} strikes the best balance in precision and recall, as we can see in the good overall positions obtained in both rankings.

%\textcolor{green}{The \emph{Metaphone} algorithm shines when precision is needed, as it is among the top 3 algorithms in the F1 classification and it does not fall to the bottom of the recall rankings as \emph{MRA} and \emph{Eudex} do.
%On the other hand, \emph{Soundex} and \emph{Roger Root} would be good choices if we want to maximize recall while not hurting precision as much as $MRA_{custom}$ or \emph{Lein}. Finally, \emph{Statcan} strikes the best balance in precision and recall, as we can see in the good overall ranking positions obtained.}

%TODO: not sure this is the best placement
%\textcolor{green}{On the other hand, a phonetic algorithm can also be considered as a locality--sensitive hashing algorithm. These are algorithms that maximize the probability of a \emph{collision} for similar items; this is, that the output for slightly different input elements is the same. In terms of phonetic algorithms, this is equivalent to maximizing the probability of two similar--sounding words having the same phonetic code. From this perspective, the average number of candidates retrieved by a phonetic algorithm is in fact directly proportional to the compression ratio of the locality--sensitive hashing it is performing.}

It is interesting to note that a phonetic algorithm can also be considered as a \emph{locality--sensitive hashing algorithm}. These are algorithms that maximize the probability of a \emph{collision} for similar items; this is, that the output for slightly different input elements is the same. In terms of phonetic algorithms, this is equivalent to maximizing the probability of two similar--sounding words having the same phonetic code. From this perspective, the average number of candidates retrieved by a phonetic algorithm is in fact directly proportional to the compression ratio of the locality--sensitive hashing it is performing.

During the error analysis, particular attention was paid to those instances from the corpora where none or just a few of the algorithms were able to provide the correct answer (in our context, a \emph{hit}).
In the case of the zero--hits list, i.e. instances for which no algorithm provided a correct answer, it contains examples such as \texttt{baddest}--\texttt{worst} or \texttt{5ayin}--\texttt{saying}, variations which do not correspond to phonetic phenomena and are thus outside the scope of this work.
A few OOV words such as \texttt{thankyou} or \texttt{sheesh} also appear in this list.
However, other interesting examples not directly supported by any phonetic algorithm can also be found, as is the case of the so--called \emph{number homophones}~\citep{thurlow2003} such as \texttt{2morrow}--\texttt{tomorrow} or \texttt{4got}--\texttt{forgot}. This is to be expected since this texting phenomenon, while also being a phonetic substitution, plays with the pronunciation of digits, and phonetic algorithms usually work only with letters.
Furthermore, these examples would be easily supported in a microtext normalization system by preprocessing the input and spelling such numbers before applying the phonetic algorithm.

%However, other interesting examples not directly supported by any phonetic algorithm can also be found, as in the case of \texttt{2morrow}--\texttt{tomorrow} or \texttt{4got}--\texttt{forgot}. This is to be expected as these algorithms mainly work with letters from the English alphabet, not numbers, thus not covering this situation. Furthermore, these examples would be easily supported in a microtext normalization system by first spelling the numbers before applying the phonetic algorithm.

On the remaining lists, it is interesting to note the presence of instances where some phonetic algorithms gave a correct answer despite them not being designed to deal with that particular case.
We can mention here those algorithms whose generated codes are shortened to a specific maximum length.
Because of this shortening, such algorithms are able to cope with any misspellings occurring in those parts of the original word which are not finally translated into the phonetic code.
For example, the MRA algorithm gives the correct answer \texttt{performance} for the input word \texttt{perfomence} as the second \texttt{r} is not encoded in \texttt{PRFMNC}.

Finally, taking into consideration the listing of non--standard orthographic forms from~\cite{thurlow2003} and the lists of errors obtained in these experiments, we can conclude the following:\footnote{These lists are available at \url{http://www.grupocole.org/software/VCS/phon}}

\begin{itemize}
	\item \emph{Shortenings} (e.g. \texttt{dec}--\texttt{december}, \texttt{epi}--\texttt{episode}) are difficult to account for as they usually remove whole chunks of characters from the original standard word, including consonants.
		The difficulty here is that consonants are the main building blocks for most phonetic codes and, consequently, important information for the algorithms has been stripped from the term to be normalized.
		This may be solved by adding an extra step in the normalization pipeline which would make use of other word similarity metrics such as the longest common subsequence, overlap coefficient or cosine distance~\citep{okazaki2010}.
		
	\item \emph{Contractions} (e.g. \texttt{frm}--\texttt{from}, \texttt{lov}--\texttt{love}) can be dealt with as they generally include the most meaningful details of the standard word, which mainly consist of its consonants.
	
	\item \emph{g--Clippings} (e.g. \texttt{losin}--\texttt{losing}, \texttt{frikin}--\texttt{freaking}) are a simple but problematic texting phenomenon for many of the algorithms studied.
		Most of them are able to incidentally handle it through their trimming of the output codes.
		In this way, in sufficiently long  words the final \texttt{g} consonant is not taken into account for the encoding, hence bypassing the need for specific rules for managing it.
		On the other hand, it is worth noting that algorithms like \emph{Beider--Morse} or \emph{Phonex}, with a wide range of encoding rules, do handle this particular scenario effectively. 

	\item Handling other types of \emph{clippings} is possible for a wide range of algorithms if they perform some kind of clipping themselves during encoding (e.g. \texttt{luk}--\texttt{luck}, \texttt{metalic}--\texttt{metallic}) or when the clipping only affects vowels or non--pronounceable characters (e.g. \texttt{ther}--\texttt{their}, \texttt{oclock}--\texttt{o'clock}). 

	\item \emph{Acronyms} and \emph{initialisms} (e.g. \texttt{omg}--\texttt{oh my god}, \texttt{lol}--\texttt{laughing out loud}) are not included in the evaluation datasets. 
		In any case, they are considered to be outside the scope of phonetic phenomena and, consequently, are not consistently supported by any phonetic algorithm.
		Nevertheless, they may be normalized using specialized dictionaries~\citep{doval2015,Han:2011:LNS:2002472.2002520}. %\footnote{\url{https://lingo2word.com/}}
		
	\item In the case of \emph{homophony} phenomena, letter homophones (e.g. \texttt{b}--\texttt{be}, \texttt{r}--\texttt{are}) are generally supported, whereas, as noted earlier, number homophones (e.g. \texttt{4got}--\texttt{forgot}, \texttt{in2}--\texttt{into}) are not. 

	\item Other misspellings, typos, non--conventional spellings and accent stylization (e.g. \texttt{acount}--\texttt{account}, \texttt{basterds}--\texttt{bastards}, \texttt{huni}--\texttt{honey}, \texttt{dat}--\texttt{that}) can be handled by most algorithms as long as the sequence of consonants was preserved in the resulting non--standard word.
		In the case of examples as \texttt{eva}--\texttt{ever} or \texttt{ova}--\texttt{over}, they are only supported by the lowest precision algorithms or by those  having specific rules for dealing with such phenomena.
\end{itemize}

%\textcolor{green}{In the final text normalization system, there will be multiple candidate generators which should complement each other. Hence, the limitations mentioned previously for phonetic algorithms will be tackled by other modules such as spell checkers or non--standard to standard text dictionaries.} 

% % % % % % % % % % % % % % % % % % % % % % % % % % % % % % % % % % % %
%
% 5. RELATED WORK
%
% % % % % % % % % % % % % % % % % % % % % % % % % % % % % % % % % % % %

\section{Related work}
\label{sect-related-work}
As already mentioned in the introductory section of this work,
%Section~\ref{sect-phonetic-algorithms}, 
phonetic algorithms have been traditionally used for personal name matching.
In the literature, the design of a new phonetic algorithm tends to be coupled with a comparative study with the contemporary state of the art in order to highlight its strengths.
However, in most cases it is not an exhaustive study, since only a small part of the existing phonetic algorithms are considered, and also because it focuses on the 
name--matching task, as in the case of the works of~\cite{hood2004}, \cite{beider2008}, \cite{mokotoff2007}, \cite{holmes2002} and \cite{parmar2014}.
Some exceptions in which a broader comparative study was performed are the studies made by~\cite{lynch1977} and~\cite{lait1996}. 

There are also purely comparative studies of name--matching algorithms where a wider range of techniques were considered, including phonetic algorithms and other types of similarity metrics. This is the case of the works of~\cite{christen2006}, \cite{snae2007}, \cite{bilenko2003}, \cite{branting2003} or \cite{galvez2006}. In their conclusions, these authors propose a series of recommendations to select the right approach or algorithm for a particular setup or domain, much in the same vein of the present work.
However, even in these cases, they only compare a small subset of the phonetic algorithms available and, more importantly, they do so in the context of a different task.

Moving on from the name--matching task, the work of~\cite{pinto2012} enters the domain of microtexts.
However, the authors only provide a comparative study between the original Soundex and their proposed improvements, which are not publicly available.
Furthermore, their case is not that of microtext normalization either.

Finally, it is worth mentioning the recent work of~\cite{fuentes2016}, which compares notably more phonetic algorithms than previous works although in a different scenario: word recognition in Spanish microtext mining.
%\textcolor{blue}{This may be the most similar work to the present contribution, although it focuses on precision and accuracy metrics; once again, it is not tailored to the microtext normalization task; and, finally, it uses Spanish microtexts instead of English ones, this being a highly language-dependant process.}
This may be the most similar work to the present contribution, although it focuses on precision and accuracy metrics, Spanish microtexts and, once again, it is not tailored to the microtext normalization task.

In general, these comparative studies mostly use precision and F1 metrics for their quantitative analysis, defining them accordingly for the task at hand.
%\cite{pinto2012}, on the other hand, use the Jaccard similarity measure on the phonetic codes for the non--standard input text and the corresponding standard text.
Some of them also give qualitative insights in order to exemplify the behaviour of the phonetic algorithms in each particular use case.
On our end, we follow the common trend of using precision and F1, while also explicitly including the recall and other interesting measures such as the average number of normalization candidates retrieved by each algorithm.
Likewise, we perform a qualitative study based on a classification of the texting phenomena.
Overall, we present a wider comparison of phonetic algorithms than in previous work, focusing on the task of microtext normalization, while using tried and tested methods for performance measurements.

% % % % % % % % % % % % % % % % % % % % % % % % % % % % % % % % % % % %
%
% 6. CONCLUSIONS AND FUTURE WORK
%
% % % % % % % % % % % % % % % % % % % % % % % % % % % % % % % % % % % %

\section{Conclusions and future work}
\label{sect-conclusions}

%\textcolor{blue}{Aiming at reducing the input noise of information processing systems working on microtexts as those used in Twitter and other microblogging services, the present work focuses on the evaluation of} a wide range of English state--of--the--art phonetic algorithms within the context of generating normalization candidates in microtext normalization tasks.
%\textcolor{blue}{The present contribution focuses on the evaluation and analysis of a wide range of English state--of--the--art phonetic algorithms within the context of generating normalization candidates for preprocessing the input of information processing systems working on microtexts as those used in Twitter and other microblogging services.}
In this work we have evaluated a wide range of English state--of--the--art phonetic algorithms within the context of generating normalization candidates in microtext normalization tasks.
This work constitutes, to the best of our knowledge, the only wide--range comparative study of its kind.
We perform both qualitative and quantitative analyses ---adapting the usual performance metrics to our domain--- in order to identify the most salient properties of these phonetic algorithms and their appropriateness for the task at hand.
We expect that the results obtained  will be of help to both developers and researchers of this field when building new intelligent systems for microtext information processing.

Seeking reproducibility and simplicity by using currently existing implementations and publicly available datasets, we have measured the performance of these algorithms, and their strengths and weaknesses were analysed to identify the best algorithms in terms of a compromise between precision and recall.
In the end, we have found that the choice of  phonetic algorithm depends heavily on the capabilities of the subsequent candidate selection mechanism to be applied within the microtext normalization pipeline.
The faster it can make the right selections among big enough input sets of candidates, the more we can sacrifice in terms of the precision of the phonetic algorithm in favour of coverage.
This would be desirable since when obtaining a low number of normalization candidates, the system would run the risk of imposing an upper limit to its overall performance at this early stage.

Finally, as future lines of work, we plan to continue working on a more developed microtext normalization system and, in this way, improve the preliminary results obtained in our previous approach in this field~\citep{doval2015}.
The results obtained in this study will greatly impact the design of the candidate selection method, which will be our main focus hereafter.
The objective will be to obtain an accurate and efficient method that would allow us to take full advantage of wide--coverage phonetic algorithms in the candidate generation step.
With respect to the limitations of the phonetic algorithms studied here when facing some texting phenomena, these can be tackled by other modules in the candidate generation step of the final normalization system. This may include, for instance, the use of spell checkers and non--standard--to--standard--text dictionaries~\citep{doval2015,Han:2011:LNS:2002472.2002520} or the integration of specialized word tokenizers~\citep{DovGomVil2016a_ENG}.
%\textcolor{green}{On the other hand, we have also noted the limitations of the phonetic algorithms studied here, which will be tackled by other modules in the candidate generation step of the final normalization system. This may include spell checkers and non--standard to standard text dictionaries.}
Extending this work to other languages such as Spanish~\citep{Alegria_etal2015a} or  even code-switching scenarios~\citep{VilAloGom2016a} is another possibility to be considered.%, always within the limits imposed by the availability of resources, specially with respect to phonetic algorithms.}

% % % % % % % % % % % % % % % % % % % % % % % % % % % % % % % % % % % %
%
% * ACKNOWLEDGEMENTS *
%
% % % % % % % % % % % % % % % % % % % % % % % % % % % % % % % % % % % %

\section*{Acknowledgements} 
{\small This research has been partially funded by the Spanish Ministry of Economy, Industry and Competitiveness (MINECO) through projects TIN2017-85160-C2-1-R, TIN2017-85160-C2-2-R, FFI2014-51978-C2-1-R and FFI2014-51978-C2-2-R, and by the Autonomous Government of Galicia through projects ED431D-2017/12, ED431B-2017/01 and ED431D~R2016/046. Moreover, Yerai Doval is funded by the Spanish State Secretariat for Research, Development and Innovation (which belongs to MINECO) and by the European Social Fund (ESF) under a FPI fellowship (BES-2015-073768) associated to project FFI2014-51978-C2-1-R.}

\end{document}